\documentclass[11pt]{article}

\usepackage{amsmath, amssymb}
\usepackage{graphicx}
\usepackage{hyperref}
\usepackage{natbib}
\usepackage{geometry}
\usepackage{setspace}
\usepackage{float}
\usepackage{booktabs}

\usepackage{titlesec}
\titlespacing*{\section}{0pt}{2.5ex}{1.5ex}

\begin{document}

\title{Valid Feature-Level Inference for Tabular Foundation Models via the Conditional Randomization Test}
\author{Mohamed Salem}
\date{}
\maketitle

\begin{abstract}
Modern machine learning models are highly expressive but notoriously difficult to analyze statistically. In particular, while black-box predictors can achieve strong empirical performance, they rarely provide valid hypothesis tests or p-values for assessing whether individual features contain information about a target variable. This article presents a practical approach to feature-level hypothesis testing that combines the Conditional Randomization Test (CRT) with TabPFN, a probabilistic foundation model for tabular data. The resulting procedure yields finite-sample valid p-values for conditional feature relevance, even in nonlinear and correlated settings, without requiring model retraining or parametric assumptions.
\end{abstract}

\section{Introduction}

Machine learning has become the dominant paradigm for predictive modeling in many applied domains, from medicine to economics to the natural sciences \citep{rudin2019stopexplainingblackbox,zhang2025}. However, as models have grown more flexible and opaque \citep{ribeiro2016whyitrustyou, rudin2019stopexplainingblackbox}, a fundamental statistical capability has been lost: the ability to produce valid p-values for questions about feature relevance.

Classical statistical models, such as linear or generalized linear models, directly yield hypothesis tests and confidence intervals as part of their standard inference framework . In contrast, modern black-box models (neural networks, ensembles, and foundation models) typically provide only predictions \citep{rudin2019stopexplainingblackbox,zhang2025}. When practitioners attempt to recover inferential statements from these models, they often rely on heuristics or asymptotic arguments that lack formal guarantees \citep{ribeiro2016whyitrustyou}. As a result, p-values derived from black-box models are frequently miscalibrated or uninterpretable.

In practice, feature selection and interpretation in machine learning commonly relies on post-hoc attribution methods, particularly Shapley values \citep{lundberg2017unified, lundberg2020local}. Derived from cooperative game theory \citep{shapley1953value}, Shapley values assign each feature a score representing its average marginal contribution to predictions across all possible feature coalitions. While Shapley values provide a theoretically principled decomposition of model predictions and have become widely adopted in applied work, they remain fundamentally descriptive rather than inferential. Shapley values quantify how much a feature contributes to a specific model's output, but they do not test whether that contribution is statistically significant, nor do they distinguish between marginal and conditional relevance. Moreover, computing exact Shapley values is computationally prohibitive for most datasets, and approximation methods such as SHAP \citep{lundberg2017unified} introduce additional sources of variability and bias that are difficult to quantify.

In particular, we would like to answer questions of the form: does a given covariate provide information about the target beyond what is already explained by the remaining variables? Answering this question rigorously requires a hypothesis test for conditional independence, rather than a measure of marginal association or model-specific attribution.

\subsection{Related Work}

Testing whether two variables are conditionally independent given a third is a foundational problem in statistics and machine learning. Classical approaches include partial correlation tests for Gaussian data \citep{anderson2003introduction}, kernel-based methods \citep{gretton_kernel_2007, zhang2011kernel}, and regression-based frameworks that test whether residuals are independent \citep{shah2020hardness}. However, these methods typically rely on linearity, Gaussianity, or large-sample asymptotics, limiting their applicability to small, nonlinear, or mixed-type tabular data.

On the other hand, heuristic feature importance measures that function with bl;ack-box type models, such as permutation importance \citep{breiman2001random}, SHAP values \citep{lundberg2017unified}, and LIME \citep{ribeiro2016whyitrustyou}, are widely used but lack formal statistical guarantees. These methods conflate marginal and conditional relevance, often producing misleading attributions in the presence of feature correlations. 

The Model-X framework, introduced by \citet{barber2015controlling} and extended by \citet{candes2017panninggoldmodelxknockoffs}, constructs knockoff variables, synthetic covariates designed to mimic the dependence structure of the original features while being independent of the response. Knockoffs enable controlled variable selection with finite-sample false discovery rate guarantees. The Conditional Randomization Test (CRT), formalized by \citet{candes2017panninggoldmodelxknockoffs} and further developed by \citet{berrett2020conditional}, generalizes this approach by directly testing conditional independence without requiring knockoff construction. Unlike asymptotic tests, the CRT provides finite-sample valid p-values under the sole assumption that the conditional distribution $p(X_j \mid X_{-j})$ can be accurately sampled.

Recent work has explored using flexible models to approximate conditional distributions for CRT. \citet{jordon2018knockoffgan} employed generative adversarial networks (GANs) to sample from $p(X_j \mid X_{-j})$. \citet{tansey2022holdout} proposed the holdout randomization test (HRT), which uses data splitting and mixture density networks to efficiently approximate $p(X_j | X_{-j})$ without refitting the predictive model for each null sample.{\citet{katsevich2022} establish fundamental optimality results for the CRT, showing that likelihood-based test statistics are optimal against point alternatives and deriving explicit connections between machine learning prediction error and asymptotic power. Their work demonstrates that the CRT's power depends directly on the quality of the model for the response $Y$ given covariates $\mathbf{X}$, providing theoretical justification for using accurate machine learning predictors to estimate the response.

Foundation models, which are pretrained on broad data distributions and applied to downstream tasks without retraining, have achieved remarkable success in natural language processing and computer vision \citep{brown2020language, dosovitskiy2020image}. TabPFN \citep{hollmann2023tabpfntransformersolvessmall} extends this paradigm to tabular data by training a transformer via in-context learning on synthetically generated datasets. Unlike traditional tabular models, TabPFN provides calibrated posterior predictive distributions in a single forward pass, making it particularly well-suited for probabilistic inference tasks. 

\subsection{Contributions}

This article describes how a valid conditional independence test can be constructed by combining the Conditional Randomization Test with TabPFN, a probabilistic transformer model for tabular data. Our work differs from prior approaches by leveraging a pretrained foundation model which performs Bayesian-style inference without task-specific training, offering both flexibility and computational efficiency. The resulting procedure produces valid p-values for feature relevance while retaining the flexibility of modern machine learning models.\footnote{
A reference implementation of the proposed TabPFN-based Conditional Randomization Test, including all experiments reported here, is available at
\url{https://github.com/msalem7777/tabpfn-crt}.
}

\section{Problem Formulation}

Let $Y$ denote a target variable and let $X = (X_1, \dots, X_p)$ denote a set of covariates. For a given feature index $j$, we consider the null hypothesis
\[
H_0: \quad Y \perp\!\!\!\perp X_j \mid X_{-j},
\]
where $X_{-j}$ denotes all covariates except $X_j$.

This hypothesis asserts that once all other covariates are known, the feature $X_j$ contains no additional information about $Y$. Rejecting this null indicates that $X_j$ is conditionally relevant to the target, while failing to reject suggests that any apparent association can be explained by correlations with other variables.

Importantly, this is a population-level statement about the joint distribution of $(X, Y)$ and does not depend on any particular predictive model.

\section{Methodology}\label{sec:method}


The Conditional Randomization Test, introduced by \citet{candes2017panninggoldmodelxknockoffs} and further developed by \citet{berrett2020conditional}, provides a general framework for testing conditional independence. The CRT proceeds by constructing a null distribution that preserves all aspects of the data-generating process except for the dependence between $X_j$ and $Y$.

Concretely, the CRT replaces the observed feature values $X_j$ with draws from their conditional distribution given the remaining covariates, $p(X_j \mid X_{-j})$. These replacements preserve the dependence structure among the covariates while breaking any direct link between $X_j$ and the target. A test statistic computed on the original data is then compared to the same statistic computed on multiple conditionally randomized datasets.

Under the null hypothesis, the observed statistic is exchangeable with its null counterparts, yielding valid p-values even in finite samples.


A central challenge in applying the CRT is accurately modeling the conditional distribution $p(X_j \mid X_{-j})$, particularly when the covariates are nonlinear, high-dimensional, or of mixed type. Traditional approaches often rely on parametric assumptions or require training separate generative models for each feature.

TabPFN \citep{hollmann2023tabpfntransformersolvessmall, hollmann_accurate_2025} offers an appealing alternative. TabPFN is a pretrained transformer that performs Bayesian-style inference over tabular datasets in a single forward pass. It can model both regression and classification problems and provides full posterior predictive distributions without task-specific retraining.

These properties make TabPFN well suited for both components of the CRT: modeling $p(Y \mid X)$ to evaluate predictive performance, and modeling $p(X_j \mid X_{-j})$ to generate conditionally valid null features.

The choice of test statistic $T(X, Y, Z)$ significantly impacts power. \citet{katsevich2022} show that likelihood-based statistics—which measure how well the data fit a learned model $\hat{f}_{Y|X,Z}$—are optimal against point alternatives. This theoretical result motivates our use of TabPFN's log-likelihood as the test statistic, as TabPFN provides calibrated posterior predictive distributions without requiring parametric assumptions. To compare the observed data with its conditionally randomized counterparts, we use the expected log predictive density (ELPD) as the test statistic. Given an evaluation dataset $\{(x_i, y_i)\}_{i=1}^n$, the statistic is defined as
\[
T_{\text{obs}} = \frac{1}{n} \sum_{i=1}^n \log p(y_i \mid x_i),
\]
where $p(y \mid x)$ is the posterior predictive distribution produced by TabPFN.

The ELPD is a proper scoring rule \citep{Gneiting01032007}, meaning that it is maximized in expectation by the true conditional distribution. As a result, it provides a principled measure of predictive quality that naturally accommodates both regression and classification settings.


To generate the null distribution, we first fit a TabPFN model to approximate the conditional distribution $p(X_j \mid X_{-j})$. For continuous features, we sample from this distribution using predicted quantiles; for categorical features, we sample from the predicted class probabilities.

For each null draw, we replace the observed values of $X_j$ with conditionally sampled values and recompute the ELPD using the same predictive model for $Y$. Repeating this procedure yields a collection of null statistics $\{T^{(1)}, \dots, T^{(B)}\}$ that represent predictive performance under the null hypothesis.


The p-value is computed as
\[
p = \frac{1 + \sum_{b=1}^B \mathbb{I}\{T^{(b)} \ge T_{\text{obs}}\}}{B + 1},
\]
which treats the observed statistic as one draw among $B+1$ exchangeable values. This construction ensures finite-sample validity and avoids degenerate p-values.



\section{Simulation and Ablation Studies}

We evaluate the proposed TabPFN-based Conditional Randomization Test (CRT) on a diverse suite of synthetic datasets. The goals of these experiments are threefold: (i) to assess finite-sample type-I error control under the null, (ii) to evaluate power for detecting conditionally relevant features across linear and nonlinear regimes, and (iii) to study the robustness of the procedure under correlated and proxy-variable structures.

Across all experiments, p-values are computed using the finite-sample CRT construction described in Section~\ref{sec:method}, with the expected log predictive density (ELPD) as the test statistic. Unless otherwise noted, each experiment uses $B=1000$ conditional resamples and is repeated over multiple random splits to reduce Monte Carlo variability.

\subsection{Synthetic Data Generating Processes}

We begin with a comprehensive set of synthetic data generators designed to isolate specific statistical challenges commonly encountered in practice. Each generator produces a feature matrix $X \in \mathbb{R}^{n \times p}$, a target $y$, and a known set of conditionally relevant features. This ground truth allows us to directly measure power and type-I error.

Each synthetic generator produces a feature matrix 
$X \in \mathbb{R}^{n \times p}$ and response $Y$.
Unless otherwise stated, noise terms are independent Gaussian:
$\epsilon \sim \mathcal{N}(0,1)$.

\subsubsection{Linear Regimes}

\paragraph{Linear Sparse.}
Features are independently generated as
\[
X_j \sim \mathcal{N}(0,1), \quad j=1,\dots,10.
\]
The response is
\[
Y = 3X_1 - 2X_2 + X_3 + \epsilon,
\]
with relevant set $\mathcal{R} = \{1,2,3\}$.
This setting evaluates detection of sparse linear signal amid irrelevant noise variables.

\paragraph{Linear Dense.}
With $p=5$ independently generated features
\[
X_j \sim \mathcal{N}(0,1),
\]
we define
\[
Y = \sum_{j=1}^{5} X_j + \epsilon,
\]
so that $\mathcal{R} = \{1,2,3,4,5\}$.
Signal is distributed uniformly across all dimensions.

\paragraph{Weak Signal.}
Under the same independent feature generation,
\[
Y = 0.5 X_1 + 0.5 X_2 + \epsilon,
\]
with $\mathcal{R} = \{1,2\}$.
This tests power in low signal-to-noise settings.

\paragraph{Noise Block.}
We generate $p=20$ independent Gaussian features and define
\[
Y = X_1 + X_2 + \epsilon,
\]
so that $\mathcal{R} = \{1,2\}$.
This evaluates robustness to high-dimensional irrelevant covariates.

\paragraph{Correlated Features.}
We generate
\[
X_1 \sim \mathcal{N}(0,1),
\quad
X_2 = X_1 + 0.1 \epsilon_2,
\]
with additional independent noise features.
The response is
\[
Y = X_1 + \epsilon.
\]
Here $X_1$ and $X_2$ are highly correlated 
($\rho \approx 0.7$), but only $X_1$ is conditionally relevant:
$\mathcal{R} = \{1\}$.
This tests whether the method distinguishes conditional relevance from marginal association.

\subsubsection{Nonlinear Regimes}

\paragraph{Friedman 1.}
Features are sampled uniformly:
\[
X_j \sim \mathrm{Uniform}(0,1).
\]
The response is
\[
Y =
10\sin(\pi X_1 X_2)
+ 20(X_3 - 0.5)^2
+ 10X_4
+ 5X_5
+ \epsilon,
\]
with $\mathcal{R} = \{1,2,3,4,5\}$.

\paragraph{Friedman 2.}
\[
Y =
X_1^2
+ X_2 X_3
- X_4
+ \sin(X_5)
+ \epsilon,
\]
with $\mathcal{R} = \{1,2,3,4,5\}$.

\paragraph{Friedman 3.}
\[
Y =
\arctan\!\left(\frac{X_1 + X_2}{X_3 + 0.1}\right)
+ X_4^2
+ \epsilon,
\]
with $\mathcal{R} = \{1,2,3,4\}$.

\paragraph{XOR Interaction.}
With Gaussian features,
\[
Y = \mathbf{1}\{X_1 > 0\}
\oplus
\mathbf{1}\{X_2 > 0\},
\]
where $\oplus$ denotes logical XOR.
Only interaction effects are present, with
$\mathcal{R} = \{1,2\}$.

\paragraph{Additive Plus Interaction.}
\[
Y = X_1 + X_2 X_3 + \epsilon,
\]
with $\mathcal{R} = \{1,2,3\}$.

\paragraph{Threshold Feature.}
With $X_j \sim \mathrm{Uniform}(0,1)$,
\[
Y = \mathbf{1}\{X_1 > 0.5\} + \epsilon,
\quad \epsilon \sim \mathcal{N}(0, 0.1^2),
\]
and $\mathcal{R} = \{1\}$.

\paragraph{Nonlinear Conditional Null.}
\[
X_2 = X_1 + 0.1\epsilon_2,
\quad
Y = \sin(X_1) + \epsilon.
\]
Although $X_2$ is marginally correlated with $Y$, 
it is conditionally independent given $X_1$.
Thus $\mathcal{R} = \{1\}$.

\subsection{Experimental Evaluation}

For each dataset in the evaluation procedure, every feature $X_j$ is tested individually under the null hypothesis that $X_j$ provides no additional information about the target $Y$ beyond the remaining covariates $X_{-j}$. Each experiment is repeated over $N_{\text{repeats}} = 5$ independent Monte Carlo runs. In every repetition, the dataset is randomly split into training (80\%) and evaluation (20\%) subsets; a TabPFN model is trained to estimate $Y \mid X$; a second TabPFN model is trained to estimate the conditional distribution $X_j \mid X_{-j}$; and a Monte Carlo null distribution is constructed using $B = 1000$ conditional resamples per feature.

For continuous features, the conditional distribution is approximated using a grid of predicted quantiles of size 
$K=200$. Null samples are generated by uniformly selecting one of the predicted quantile levels for each observation, yielding a discretized inverse-CDF approximation of the conditional distribution.

The test statistic is defined as the average predictive log-likelihood on the evaluation set, and the p-value is computed as a right-tailed Monte Carlo estimate comparing the observed statistic to its null distribution. This procedure yields feature-level p-values and rejection decisions for every feature across all repetitions and datasets.

We compute empirical power and type-I error at the dataset level. Power is defined as the proportion of truly relevant features correctly rejected, and type-I error as the proportion of irrelevant features incorrectly rejected. These quantities are averaged across Monte Carlo repetitions. 


To evaluate finite-sample calibration, we aggregate p-values across datasets and repetitions. Empirical cumulative distribution functions (ECDFs) are constructed separately for relevant and irrelevant features. Under the null, p-values for irrelevant features closely follow the Uniform$(0,1)$ distribution, indicating proper type-I error control, while p-values for relevant features concentrate near zero, demonstrating strong discriminative power. Quantile–quantile (QQ) plots further confirm stable null calibration across linear, nonlinear, interaction-based, and correlated regimes.

\begin{table}[H]
\centering
\begin{tabular}{lcccc}
\toprule
Dataset & $p$ & $|\mathcal{R}|$ & Power & Type-I Error \\
\midrule
Linear (sparse) & 10 & 3 & 1.00 & 0.03 \\
Linear (dense) & 5 & 5 & 1.00 & 0.00 \\
Weak signal & 5 & 2 & 1.00 & 0.07 \\
Noise block & 20 & 2 & 1.00 & 0.03 \\
Correlated linear & 5 & 1 & 1.00 & 0.10 \\
Friedman 1 & 10 & 5 & 1.00 & 0.04 \\
Friedman 2 & 10 & 5 & 0.60 & 0.00 \\
Friedman 3 & 10 & 4 & 0.40 & 0.00 \\
XOR interaction & 5 & 2 & 1.00 & 0.00 \\
Threshold feature & 5 & 1 & 1.00 & 0.00 \\
Conditional null & 2 & 1 & 0.00 & 0.00 \\
\bottomrule
\end{tabular}
\caption{Average empirical power and type-I error of the TabPFN Conditional Randomization Test across synthetic datasets. Results are averaged over $N_{\text{repeats}} = 5$ independent Monte Carlo runs. $p$ denotes the total number of features and $|\mathcal{R}|$ the number of conditionally relevant features.}
\label{tab:power_type1}
\end{table}

\begin{figure}[H]
\centering
\includegraphics[width=0.75\linewidth]{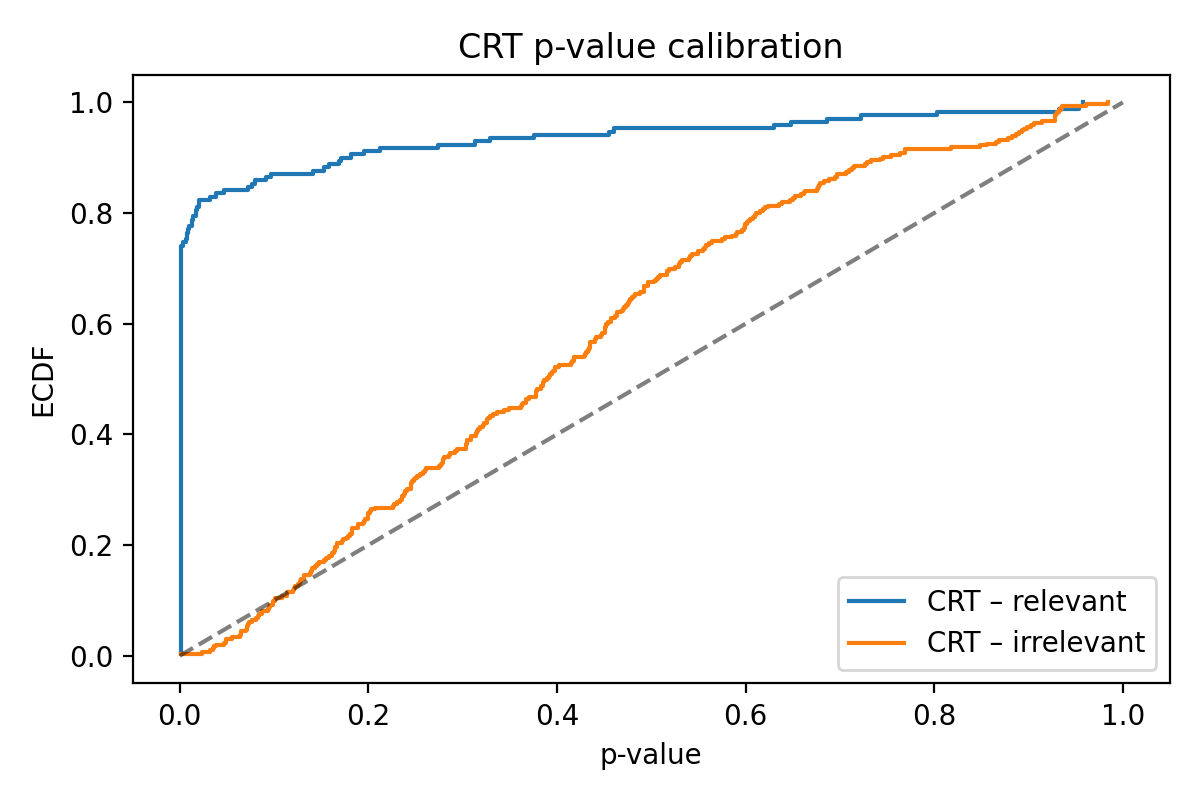}
\caption{
Empirical cumulative distribution functions of CRT p-values for conditionally relevant and irrelevant features.
Null p-values closely follow the Uniform$(0,1)$ distribution, while relevant features exhibit strong concentration near zero, indicating both valid calibration and high power.
}
\label{fig:ecdf}
\end{figure}

\begin{figure}[H]
\centering
\includegraphics[width=0.55\linewidth]{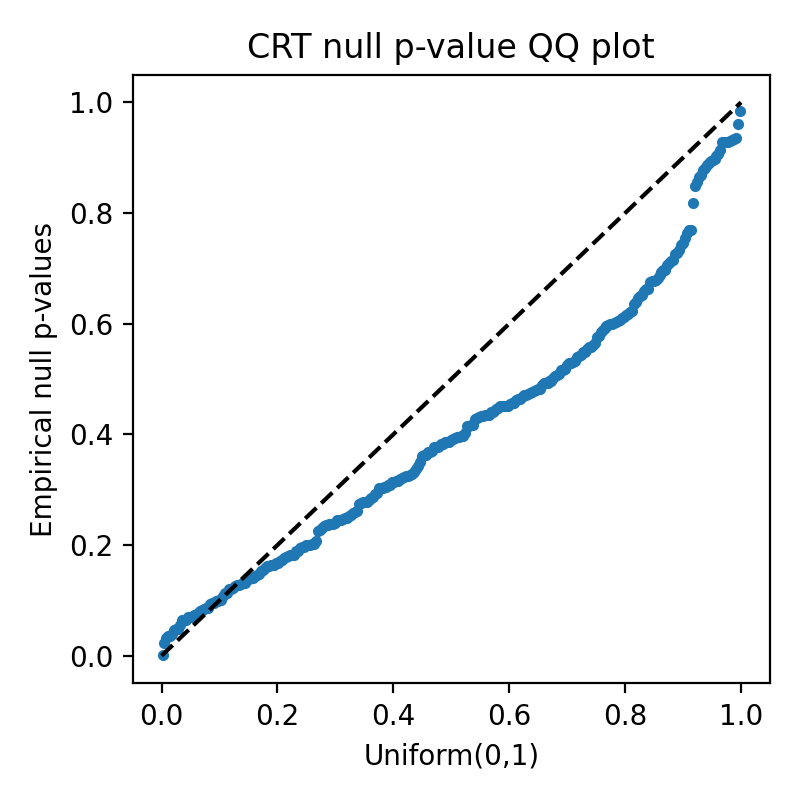}
\caption{
Quantile–quantile plot of empirical null p-values versus the Uniform$(0,1)$ distribution.
The close alignment with the diagonal is consistent with finite-sample calibration in these runs of the TabPFN-based Conditional Randomization Test across heterogeneous datasets.
}
\label{fig:qq}
\end{figure}

\subsection{Results and Interpretation}

Table \ref{tab:power_type1} reports empirical power and type-I error for the TabPFN-based CRT across 11 synthetic benchmarks, averaged over $N_{\text{repeats}} = 5$ independent Monte Carlo repetitions. Each repetition involves a random train-evaluation split, ensuring that the reported metrics reflect both algorithmic performance and sampling variability.

Across most datasets, empirical type-I error remains close to or below the nominal level $\alpha = 0.05$, with six benchmarks achieving type-I error $\le 0.03$. This indicates strong finite-sample calibration under diverse data-generating processes, including linear, nonlinear, and interaction-based regimes.

However, several datasets exhibit moderately elevated type-I error: Correlated Linear at 0.10 and Weak Signal at 0.07. These deviations suggest that the quality of the conditional model $p(X_j \mid X_{-j})$ affects null calibration, particularly in settings with sparse signal or complex functional forms. When TabPFN's approximation of the conditional distribution is imperfect, the exchangeability assumption underlying the CRT may be violated, leading to inflated false positive rates. In terms of power, the method demonstrates consistently high power across most benchmarks, achieving perfect detection (power = 1.00) in 8 of 11 datasets. Reduced power is observed in Friedman 2 (0.6), Friedman 3 (0.4), and nonlinear Conditional Null (0.00). All of which involve nonlinear forms with some interaction form between the covariates.

Figure~\ref{fig:ecdf} displays the empirical cumulative distribution functions of p-values for relevant and irrelevant features, aggregated across all datasets and repetitions. P-values for irrelevant features closely track the Uniform$(0,1)$ distribution—indicated by alignment with the diagonal—while p-values for relevant features concentrate sharply near zero. This visual diagnostic confirms both valid type-I error control and strong discriminative power. Figure~\ref{fig:qq} presents a quantile-quantile plot comparing empirical null p-values to the theoretical Uniform$(0,1)$ quantiles. The close alignment with the diagonal across the full range of quantiles provides further evidence of finite-sample calibration. Minor deviations in the upper tail are consistent with the moderately elevated type-I error observed in a subset of benchmarks.

\subsection{Limitations and Practical Considerations}

While the proposed method demonstrates strong empirical performance, several limitations and practical considerations warrant discussion.

\paragraph{Dependence on Conditional Modeling Quality.}
The validity of the CRT fundamentally relies on the ability to accurately sample from $p(X_j \mid X_{-j})$. When TabPFN poorly approximates this conditional distribution the resulting p-values may be miscalibrated. The elevated type-I error in Linear (sparse) and Friedman 3 suggests that approximation error occasionally affects null calibration, particularly in high-dimensional or complex nonlinear settings. However, this can be remedied via extended training of the model on a larger, more encompassing set of synthetic data.


\paragraph{Computational Cost.}
Each feature test requires fitting two TabPFN models (one for $Y \mid X$, one for $X_j \mid X_{-j}$) and generating $B$ conditional resamples with repeated predictions. We can alleviate the cost for multiple feature testing by calling the former once. For a dataset with $p$ features, this yields $\mathcal{O}((n^{2}+p^{2}) \cdot (B+1))$ model evaluations per repetition. While TabPFN's single-pass inference is faster than iterative retraining of traditional models, testing many features remains computationally nontrivial. For datasets with large $p$ or $n$, practitioners may need to prioritize feature subsets or use parallel computation.


\section{Discussion and Conclusion}

The combination of the Conditional Randomization Test with TabPFN provides a rare synthesis of modern machine learning flexibility and classical inferential guarantees. Unlike heuristic feature attribution methods, the resulting p-values have a clear interpretation grounded in a well-defined null hypothesis: that a feature provides no information about the target beyond what is already explained by the remaining covariates. At the same time, the approach avoids restrictive parametric assumptions and accommodates complex, correlated, and nonlinear dependencies.


This framework is particularly well suited to small and medium-sized tabular datasets, where overfitting and correlation structure often invalidate traditional importance measures. While the method depends on accurately modeling the conditional distribution of the covariates, TabPFN provides a strong practical approximation in many settings \citep{hollmann2023tabpfntransformersolvessmall, hollmann_accurate_2025}. The empirical results presented here demonstrate that this approximation is sufficient to achieve both valid type-I error control and high power across diverse synthetic benchmarks.


By leveraging TabPFN as this probabilistic engine, we obtain a practical procedure for feature-level hypothesis testing that produces finite-sample valid p-values without sacrificing modeling power. The method correctly distinguishes conditional relevance from marginal association, addressing a key limitation of existing explainability tools. Moreover, because TabPFN requires no task-specific retraining, the procedure is computationally efficient relative to approaches that rely on iterative model fitting or adversarial training.

Looking forward, several directions remain for future work. First, extending the framework to handle very large datasets or high-dimensional feature spaces will require alternative conditional samplers or approximations that scale beyond TabPFN's current design constraints. Second, integrating the CRT with causal inference frameworks, such as directed acyclic graphs or structural equation models, could enable principled causal claims rather than purely associative statements. Finally, developing diagnostics to detect when conditional modeling quality is insufficient would allow practitioners to assess the reliability of reported p-values in real-world applications.

In conclusion, this work demonstrates that modern foundation models can be integrated into rigorous statistical frameworks, offering inference without sacrificing flexibility. As machine learning systems are increasingly deployed in high-stakes domains, methods that combine predictive power with interpretable, valid inference will become essential tools for responsible data science.

\bibliographystyle{plainnat}
\bibliography{references}

\end{document}